\documentclass[final]{cvpr}

\usepackage{times}
\usepackage{epsfig}
\usepackage{graphicx}
\usepackage{amsmath}
\usepackage{amssymb}

\usepackage{xcolor}
\usepackage{multirow}
\usepackage{array}

\usepackage[pagebackref=true,breaklinks=true,colorlinks,bookmarks=false]{hyperref}

\def\cvprPaperID{3630} 
\def\confYear{CVPR 2021}
\begin{document}

\title{Probabilistic Modeling of Semantic Ambiguity for Scene Graph Generation}

\author{
    Gengcong Yang$^1$\thanks{Equal contribution.} \thanks{Work mainly done during an internship at Tencent AI Lab.},\quad Jingyi Zhang$^{2 *}$,\quad Yong Zhang$^3$\thanks{Yong Zhang, Baoyuan Wu and Yujiu Yang are the corresponding authors.},\quad Baoyuan Wu$^{4,5 \ddagger}$,\quad Yujiu Yang$^{1 \ddagger}$\\
    {\small $^1$Tsinghua Shenzhen International Graduate School, Tsinghua University,\quad $^2$University of Electronic Science and Technology of China,} \\
    {\small $^3$Tencent AI Lab,\quad $^4$School of Data Science, The Chinese University of Hong Kong, Shenzhen,} \\
    {\small $^5$Secure Computing Lab of Big Data, Shenzhen Research Institute of Big Data}\\
    {\tt\small ygc19@mails.tsinghua.edu.cn,\quad jgg\_jingyizhang@foxmail.com,\quad zhangyong201303@gmail.com}\\
    {\tt\small wubaoyuan@cuhk.edu.cn,\quad yang.yujiu@sz.tsinghua.edu.cn}
}

\maketitle

\newcommand{\oi}{\boldsymbol{o}_i}
\newcommand{\bi}{\boldsymbol{b}_i}
\newcommand{\si}{\boldsymbol{s}_i}
\renewcommand{\xi}{\boldsymbol{x}_i}
\newcommand{\xj}{\boldsymbol{x}_j}
\newcommand{\cij}{\boldsymbol{c}_{ij}}
\newcommand{\rij}{\boldsymbol{r}_{ij}}
\newcommand{\uij}{\boldsymbol{u}_{ij}}
\newcommand{\eij}{\boldsymbol{e}_{ij}}
\newcommand{\zij}{\boldsymbol{z}_{ij}}
\newcommand{\muij}{\boldsymbol{\mu}_{ij}}
\newcommand{\sigij}{\boldsymbol{\sigma}^2_{ij}}
\newcommand{\todo}[1]{\textcolor{red}{TO DO \textsf{#1}}}
\newcommand{\up}[1]{\textbf{ $\uparrow$ #1}}
\newcommand{\down}[1]{ $\downarrow$ #1}
\newcommand{\modify}[1]{\textcolor{red}{#1}}

\begin{abstract}
To generate ``accurate" scene graphs, almost all existing methods predict pairwise relationships in a deterministic manner. 
However, we argue that visual relationships are often semantically ambiguous.
Specifically, inspired by linguistic knowledge, we classify the ambiguity into three types: Synonymy Ambiguity, Hyponymy Ambiguity, and Multi-view Ambiguity.
The ambiguity naturally leads to the issue of \emph{implicit multi-label}, motivating the need for diverse predictions.
In this work, we propose a novel plug-and-play Probabilistic Uncertainty Modeling (PUM) module.
It models each union region as a Gaussian distribution, whose variance measures the uncertainty of the corresponding visual content.
Compared to the conventional deterministic methods, such uncertainty modeling brings stochasticity of feature representation, which naturally enables diverse predictions.
As a byproduct, PUM also manages to cover more fine-grained relationships and thus alleviates the issue of bias towards frequent relationships.
Extensive experiments on the large-scale Visual Genome benchmark show that combining PUM with newly proposed ResCAGCN can achieve state-of-the-art performances, especially under the mean recall metric.
Furthermore, we prove the universal effectiveness of PUM by plugging it into some existing models and provide insightful analysis of its ability to generate diverse yet plausible visual relationships.

\end{abstract}


\section{Introduction}

Scene graph generation (SGG) has been an important task in computer vision, serving as an intermediate task to bridge the gap between upstream object detection~\cite{faster_rcnn}
and downstream high-level visual understanding tasks, such as image captioning~\cite{img_cap_auto, img_cap_compre} and visual question answering~\cite{vqa_sg}.
Intuitively, the latter would get greater benefit from more human-like scene graphs.

\begin{figure}[t]
    \centering\includegraphics[width=0.47\textwidth, keepaspectratio]{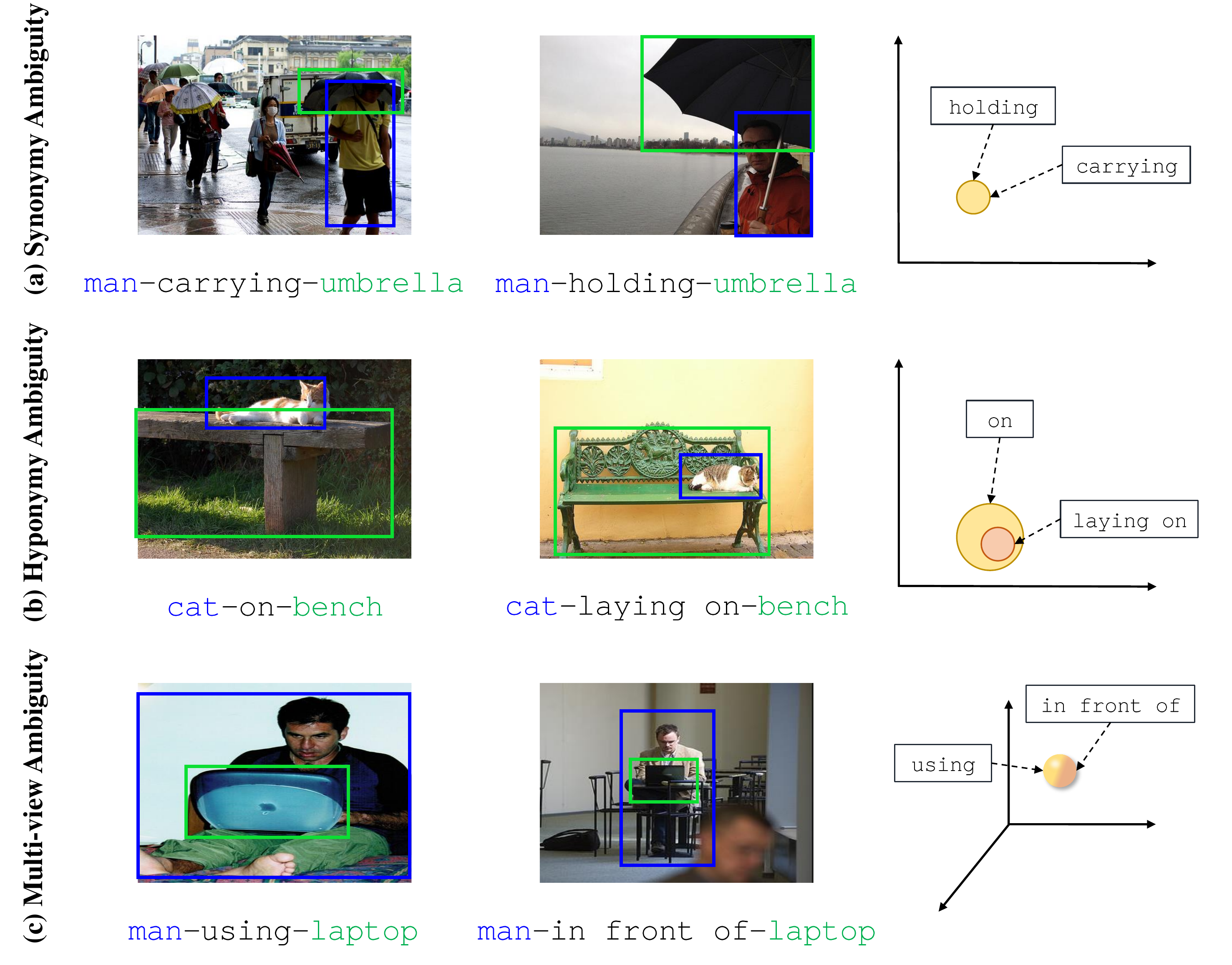}
	\caption{Examples of semantic ambiguity in Visual Genome dataset.
	The first two columns show the comparisons between two plausible predicates for similar visual scenes and the rightmost column illustrates the corresponding phenomenons in semantic space.
	(a) \texttt{carrying} and \texttt{holding} share overlapping definitions and are interchangeable to describe the relationship between a man and an umbrella.
	(b) Both \texttt{on} and \texttt{laying on} are reasonable to describe the scene where a cat is on top o a bench, despite their semantic specificity difference.
	(c) Different human annotators focus on different points of view, \ie \texttt{using} (\emph{actional}) vs. \texttt{in front of} (\emph{spatial}), to describe a working man and a laptop.}
	\label{fig:ambiguous_examples}
\end{figure}

Almost all existing works~\cite{imp, motif, kern, vctree} view SGG as an objective task and predict pairwise relationships in a deterministic manner.
Namely, given a pair of objects, these models always generate an identical predicate.
However, compared with human annotators, such methods pursue ``accurate" scene graphs but overlook the intrinsic semantic ambiguity of visual relationships.
Specifically, the collaborative annotations from human annotators tend to be diverse, covering different descriptions of relationships for similar visual scenes.

We observe that there exist multiple types of semantic ambiguity in the large-scale Visual Genome dataset.
Inspired by linguistic knowledge, we classify the ambiguity into three types.
The first type is \textbf{Synonymy Ambiguity}, where multiple synonymous predicates that share overlapping definitions are suitable to describe similar visual scenes.
For example, in Figure \ref{fig:ambiguous_examples} (a), \texttt{carrying} and \texttt{holding} are interchangeable to describe the relationship between a man and an umbrella.
If we visualize these two words in the semantic space, they should point to the same position where the visual relationship lies.
The second one is \textbf{Hyponymy Ambiguity}, indicating that different human annotators tend to use predicates across adjacent abstract levels.
One would simply use \texttt{on} to describe the visual scene where a cat is on top of a bench, while others may choose to use more fine-grained \texttt{laying on}, as shown in Figure \ref{fig:ambiguous_examples} (b).
In this case, \texttt{laying on} is a hyponym of \texttt{on}.
Namely, the semantic range of the former is included by that of the latter.
As for the third type of ambiguity, we notice that different human annotators often focus on different types of visual relationships, which originate from different points of view.
Therefore, we refer to this phenomenon as \textbf{Multi-view Ambiguity}.
An example is illustrated in Figure \ref{fig:ambiguous_examples} (c), where both \texttt{using} (\emph{actional}) and \texttt{in front of} (\emph{spatial}) are plausible to describe the relationship between a working man and a laptop.
If we abstract the visual scene in the three-dimension space, it can be a multicolor sphere that reflects different colors from different views.
Although most predicates are single-labeled in the dataset, due to the ubiquitous semantic ambiguity mentioned above, we argue that many of them should be multi-labeled, since similar visual scenes are annotated as different predicates.
We refer to the issue as an \emph{implicit multi-label} problem, which motivates the need to generate diverse predictions for visual relationships.

In this work, we focus on modeling the semantic ambiguity of visual relationships and propose a novel plug-and-play Probabilistic Uncertainty Modeling (PUM) module which can be easily deployed in any existing SGG model.
Specifically, we utilize a probability distribution to represent each union region, rather than a deterministic feature vector as in previous methods.
From a geometric perspective, the probabilistic representation allows us to map each visual relationship to a soft region in space, instead of merely a single point~\cite{word_gauss}.
For ease of modeling, we adopt Gaussian distributions to represent them.
Namely, each union region is now parametrized by a mean and variance.
The former acts like the normal feature vector as in the conventional model, whereas the latter measures the feature uncertainty.
To some extent, in this way, the feature instance of each union region can be viewed as a random variable drawn from a Gaussian distribution.
Thanks to this uncertainty modeling, ambiguous union regions will be assigned to Gaussian distributions with large variances, which generate diverse samples and result in diverse predictions.
As a byproduct, we find that PUM also manages to cover more fine-grained relationships and thus well alleviates the infamous issue of bias towards frequent relationships~\cite{kern,vctree}.

We firstly demonstrate the effectiveness of PUM on the large-scale Visual Genome benchmark.
Combining with the newly proposed Residual Cross-attention Graph Convolutional Network (ResCAGCN) in concurrent work~\cite{sgg_dual_resgcn}, we achieve state-of-the-art performances under the existing evaluation metrics, especially the mean recall metric.
Note that our PUM can serve as a plug-and-play component.
Therefore, we plug PUM into some state-of-the-art SGG models and observe obvious universal improvement over these baselines, which mainly lies in the mean recall metric again.
We owe the performance gain in the mean recall metric to the ability to generate relationships diversely, which improves the chances to hit the ground-truth with infrequent predicate labels.
We further propose oracle recall as an indirect evaluation metric to measure the diversity of multiple inferences, which takes results of multiple consecutive predictions as an ensemble and computes recall.
The oracle recall of the proposed model increases with the number of predictions, indicating that the model generates plausible relationships diversely and thus gradually covers the ground-truth more and more.

Overall, the main contributions of this work can be summarized as follows:

\begin{itemize}
    \item We notice the semantic ambiguity of visual relationships and propose a novel plug-and-play Probabilistic Uncertainty Modeling (PUM) module, which utilizes a probability distribution to represent each union region instead of a deterministic feature vector.
    \item Combining PUM with ResCAGCN, we achieve state-of-the-art performances on the large-scale Visual Genome benchmark under the existing evaluation metrics, especially the mean recall metric.
    \item Extensive evaluations demonstrate the superiority of PUM to alleviate the bias towards frequent categories when plugged into the existing SGG models, reflected in the improvement on the mean recall metric.
    \item To the best of our knowledge, we are the first to explore diverse predictions for SGG.
    We conduct experiments both qualitatively and quantitatively to demonstrate that the proposed PUM module can generate diverse yet plausible relationships.
\end{itemize}

\begin{figure*}[ht]
    \centering\includegraphics[width=\textwidth, keepaspectratio]{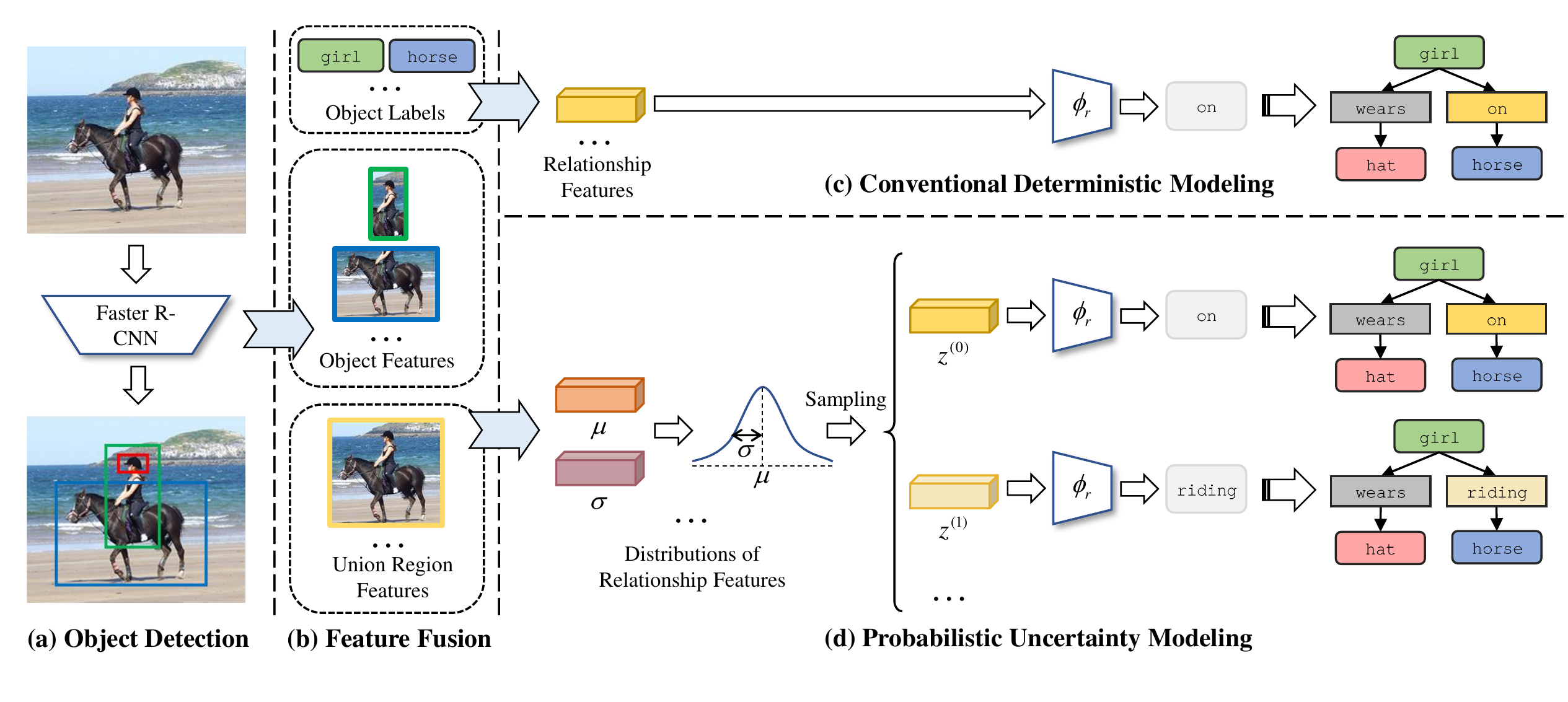}
	\caption{
	Existing SGG framework usually includes the following steps:
	(a) utilizing Faster R-CNN to obtain object proposals;
	(b) fusing features globally to obtain object labels, object features, and union region features;
	(c) conditioned on the results of the previous step, modeling each union region as a deterministic vector to predict the relationship.
	In this work, we replace (c) with Probabilistic Uncertainty Modeling (PUM) in (d), where each union region is represented by a probability distribution instead.
	In such way, diversity in scene graph generation is naturally achieved.
	}
	\label{fig:overview}
\end{figure*}

\section{Related Work}

\noindent \textbf{Scene Graph Generation.}
Visual relationships have raised wide concern in the computer vision community since Lu \etal~\cite{vrd_lu} formalized Visual Relationship Detection as a visual task.
In many early works~\cite{vrd_lu,vrd_context_inter,vrd_deepstruct,vrd_relpn,vrd_vtranse}, objects and pairwise relationships were detected independently.
Such models overlooked the rich visual context and led to suboptimal performance.
To make full use of visual context, later methods took account of the whole image and resorted to various kinds of message passing mechanisms~\cite{imp,sgg_fac_net,sgg_from_opc,sgg_graph_rcnn,motif,vctree,sgg_img_rec,sgg_vis_pat,sgg_gps}.
For example, Xu \etal~\cite{imp} were the first to formally define the problem of Scene Graph Generation and addressed it with iterative refinement via message passing.
Afterward, Zellers \etal~\cite{motif} represented the global context via LSTM, a kind of recurrent sequential architecture.
More recently, Chen \etal~\cite{kern} incorporated statistical correlations into the graph propagation network.
Meanwhile, Tang \etal~\cite{vctree} composed dynamic tree structures to allow content-specific message passing.
While all these methods overlook the semantic ambiguity of visual relationships and make inferences in a deterministic manner, we propose to address the ambiguity via probabilistic modeling.

\noindent \textbf{Uncertainty Modeling.}
Conventionally, the high-level representation of an input instance, \eg an image or a word, is modeled as a fixed-length feature vector, namely, a single \emph{point} in $\mathbb{R}^D$.
However, such a point estimate is not sufficient to express uncertainty.
In recent years, Gaussian embedding has been getting more attention in deep learning since \cite{word_gauss} utilized it to represent words instead of the conventional word2vec~\cite{word_vec}, where the covariance naturally measures the ambiguity of the words.
In the computer vision community, there exist prior works on modeling images as Gaussian distributions~\cite{hedged_embed,robust_reid,face_uncer}.
However, all the existing SGG methods represent each union region as a deterministic vector.
In this work, we are the first to focus on the intrinsic semantic ambiguity of visual relationships and model each union region as a Gaussian distribution.

\noindent \textbf{Diverse Predictions.}
Generally, there are two types of approaches to generate multiple diverse predictions.
One is to train multiple models and aggregate their predictions.
To better obtain diversity in the union of predictions, Multiple Choice Learning (MCL)~\cite{mcl} and other variants~\cite{smcl, cmcl, vmcl} were proposed to establish cooperation among all the models and train each to specialize on one particular subset of the data distribution.

Another is to infer diverse predictions from a single model.
Before deep learning, this type of methods mainly focused on probabilistic graphical models~\cite{m_best}.
Afterward, existing single-model methods can be roughly categorized into two types: 1) via random noise added to Generative Adversarial Networks (GANs), applied in image captioning~\cite{img_cap_diverse}, image annotation~\cite{tagging_diverse}, text generation~\cite{txt_gen_diverse} and so on; 2) mapping an instance to a probability distribution in latent space and sampling from it~\cite{vqg_diverse, dialog_vae1, dialog_vae2}.
Our model can be regarded as the second category.
While the other methods in this category usually utilize Variational Autoencoders (VAEs) in a generative way,
we simply make use of Gaussian distributions to generate stochastic representations without a reconstruction loss and achieve diversity via this stochasticity in the inference stage.

\section{Approach}

Generally, a \emph{scene graph} is a structured representation that describes the contents of a visual scene, which encodes object instances via nodes and relationships between objects via edges~\cite{scene_graphs}.
As defined by Xu \etal~\cite{imp}, the task of Scene Graph Generation is to generate an accurate visually-grounded scene graph associated with an image.

Mathematically, a scene graph can be defined as $G = \{B, O, R\}$, where $B$ is a set of bounding boxes, $O$ denotes object labels, and $R$ denotes relationship labels.
Conventionally, given an image $I$, the probability distribution of a scene graph $P(G|I)$ is decomposed into three factors~\cite{motif,kern}:
\begin{equation}
    P(G|I) = P(B|I) P(O|B, I)P(R|O, B, I).
\end{equation}
Firstly, the widely used Faster R-CNN~\cite{faster_rcnn} is utilized to model $P(B|I)$ and generates a set of object proposals.
Next, conditioning on the candidate bounding boxes, the object model $P(O|B, I)$ predicts the class label regarding each box.
Finally, based on the result of object detection, the relationship model $P(R|O, B, I)$ infers the relationship of each object pair, leading to the whole scene graph for the current image.
Existing works treat $P(R|O, B, I)$ as a deterministic model, which always generates an identical label for the same object pair.
This framework is illustrated in Figure \ref{fig:overview} (a)(b)(c).
However, such a method overlooks the intrinsic semantic ambiguity of visual relationships and is likely to get stuck in the issue of biased predictions~\cite{kern,vctree}, with the tendency to generate frequent and ``safe" labels.

In this work, we propose a plug-and-play module for the relationship model, named Probabilistic Uncertainty Modeling (PUM), which addresses the semantic ambiguity mentioned above in a probabilistic manner.
We replace the conventional deterministic modeling with PUM, as illustrated in Figure \ref{fig:overview} (d).
To better demonstrate the effectiveness of PUM, we adopt the newly proposed ResCAGCN~\cite{sgg_dual_resgcn} as our object model, which is introduced in Section \ref{sec:obj_model}.
However, note that any object model from the existing SGG methods will be compatible with PUM theoretically. 
Then, we describe our PUM module in detail in Section \ref{sec:uncer_modeling}.

\subsection{Object Model} 
\label{sec:obj_model}

In our approach, we take Residual Cross-attention Graph Convolutional Network (ResCAGCN) from \cite{sgg_dual_resgcn} as our object model to fuse object features and predict object labels.

The core of ResCAGCN is the cross-attention module ($\mathcal{CA}$), which is designed to capture the semantic relevance among the object features and the pairwise union region features.
The module is formulated as:
\begin{equation}
    \begin{aligned}
    \mathcal{CA} \left(\boldsymbol{x}_i, \boldsymbol{x}_j\right)
    & = \left(\boldsymbol{W}_{i} \boldsymbol{x}_i \odot \sigma\left(\boldsymbol{W}_{j}^{'} \boldsymbol{x}_j\right) \oplus\boldsymbol{W}_{i} \boldsymbol{x}_i\right) \\
    & \odot\left(\boldsymbol{W}_{j} \boldsymbol{x}_j \odot \sigma\left(\boldsymbol{W}_{i}^{'} \boldsymbol{x}_i\right) \oplus \boldsymbol{W}_{j} \boldsymbol{x}_j\right),
    \end{aligned}
\end{equation}
where $\odot$ and $\oplus$ denote element-wise product and sum, respectively.
$\sigma$ is the sigmoid function to normalize the attention scores.
All $\boldsymbol{W}_{*}$ denote linear transformations to embed features into the same dimension, both here and below.

Given two object features $\xi$ and $\xj$, and their union region feature $\uij$, in order to model the contextual information, ResCAGCN utilizes the cross-attention module to compute the contextual coefficient $\cij$, which is formulated as:
\begin{equation}
    \boldsymbol{c}_{i j}=\sigma\left(\boldsymbol{W}_c^{T}\left(\mathcal{C} \mathcal{A}\left(\mathcal{C} \mathcal{A}\left(\boldsymbol{x}_{i}, \boldsymbol{x}_{j}\right), \boldsymbol{u}_{i j}\right)\right)\right).
\end{equation}

Instead of directly using the aggregated features as the output features, ResCAGCN uses a residual connection to add them back to the original features:
\begin{equation}
    \hat{\boldsymbol{x}}_{i} =
    \boldsymbol{x}_{i} + 
    \operatorname{ReLU}\left(\boldsymbol{W}_{1} L N\left(\boldsymbol{W}_{2} \sum_{j \in \mathcal{N}_{i}} \boldsymbol{c}_{i j} \otimes \boldsymbol{W}_{3} \boldsymbol{x}_{j}\right)\right),
\end{equation}
where $\otimes$ denotes Kronecker product,
$\mathcal{N}_{i}$ denotes the $i$-th node's neighborhood,
and $L N$ denotes layer normalization~\cite{layer_norm}.
The refined object features $\hat{\boldsymbol{x}}_{i}$ are then fed into a classifier to predict the object labels.

\subsection{Probabilistic Uncertainty Modeling} \label{sec:uncer_modeling}

Conventionally, the union of two proposals is represented as a single point in space, namely, \emph{point embedding}~\cite{hedged_embed}.
As \cite{word_gauss} observed, however, such point estimate does not naturally express the uncertainty induced by the input.
In the case of visual relationships, this could be caused by ambiguous annotations, \eg \texttt{holding} and \texttt{looking at} may be both plausible to describe a scene containing a man and a cell phone.

As shown in Figure \ref{fig:overview} (d), in order to capture the intrinsic uncertainty of visual relationships, we propose to \emph{explicitly} model the feature distribution of each union region as Gaussian. 
That is, we represent each union region as \emph{stochastic embedding} instead of the conventional \emph{point embedding}.
From a stochastic perspective, the final representation of each union region is no longer a deterministic vector but randomly drawn from a Gaussian distribution.
As a result, we can generate predicates diversely for the same object pair, leading to diversity in scene graph generation.

\noindent \textbf{Stochastic Representation.}
For each object pair, following ResCAGCN, we first fuse their contextual object features $\hat{\boldsymbol{x}}_i$ and $\hat{\boldsymbol{x}}_j$, as described in Section \ref{sec:obj_model}, and their visual union region feature $\uij$ to obtain relationship feature:
\begin{equation} \label{eq:rel_feature}
\boldsymbol{e}_{i j}=\hat{\boldsymbol{x}}_{i} \diamond \hat{\boldsymbol{x}}_{j} \diamond \boldsymbol{u}_{i j},
\end{equation}
where $\diamond$ denotes the fusion function defined in \cite{vqa_count,vctree},
$\boldsymbol{x} \diamond \boldsymbol{y}=\operatorname{ReLU}\left(\boldsymbol{W}_{x} \boldsymbol{x}+\boldsymbol{W}_{y} \boldsymbol{y}\right)-\left(\boldsymbol{W}_{x} \boldsymbol{x}-\boldsymbol{W}_{y} \boldsymbol{y}\right)^2$.
Based on each fused relationship feature, we define the associated representation $\zij$ in latent space as a Gaussian distribution,
\begin{equation}
    p(\zij|\eij) = \mathcal{N}(\zij; \muij, \sigij),
\end{equation}
where $\muij$ and $\sigij$ refer to mean vector and diagonal covariance matrix respectively.
They are formulated as:
\begin{equation} \label{eq:mu}
    \muij = \boldsymbol{W}_\mu \eij,
\end{equation}
\begin{equation} \label{eq:sig}
    \sigij = \boldsymbol{W}_\sigma \eij.
\end{equation}

At test time, we sample multiple $\zij$s, feed them into the relationship classifier $\phi_r$ respectively and compute the average posterior probability distribution:
\begin{equation} \label{eq:rel_cls}
    P_{ij} = \frac {1}{K} \sum^K_{k=1} \phi_r (\zij^{(k)}),
\end{equation}
where $\zij^{(k)} \sim \mathcal{N}(\muij, \sigij)$, and $K$ is the number of samples drawn from the Gaussian.
Then we simply take the argmax of $P_{ij}$ as the predicted relationship label.

\noindent \textbf{Uncertainty-aware Loss.}
$\muij$ can be viewed as the original deterministic representation of the union box, while the random variable $\zij$ serves as a stochastic representation sample.
Here, we consider both representations and feed them into $\phi_r$ respectively.
Then, we train the relationship model with cross-entropy loss,
\begin{equation}
    L_{ce} = (1 - \lambda) \mathbf{CE}(\phi_r(\boldsymbol{\mu}), \boldsymbol{y}) + \lambda (\mathbb{E}_{z \sim p(z|e)} \mathbf{CE}(\phi_r(\boldsymbol{z}), \boldsymbol{y})),
\end{equation}
where $\lambda$ is the weight to trade off between deterministic prediction and stochastic predictions, and $\mathbf{CE}$ means cross-entropy loss.
Note that we omit the subscripts $ij$ for clarity.

In practice, we approximate the expectation term via Monte-Carlo sampling from $\boldsymbol{z}^{(k)} \sim p(\boldsymbol{z}|\boldsymbol{e})$:
\begin{equation} \label{eq:ce_loss}
    L_{ce} \approx (1 - \lambda) \mathbf{CE}(\phi_r(\boldsymbol{\mu}), \boldsymbol{y}) + \lambda (\frac {1}{N} \sum^N_{k=1} \mathbf{CE}(\phi_r(\boldsymbol{z}^{(k)}), \boldsymbol{y})),
\end{equation}
where $N$ is the number of samples drawn from the Gaussian.
It is clear that conventional deterministic training can be seen as a special case of Eq. \ref{eq:ce_loss} where $\lambda$ is set to 0.

Inspired by \cite{robust_reid}, as training progresses, the variance $\boldsymbol{\sigma}^2$ always decreases with $L_{ce}$ alone and reverts our stochastic representation back to conventional deterministic model.
This problem could be alleviated by the following regularization term:
\begin{equation} \label{eq:reg_term}
    L_{reg} = \max (0, \gamma - h(\mathcal{N}(\boldsymbol{\mu}, \boldsymbol{\sigma}^2))),
\end{equation}
where $\gamma$ is a margin to bound the uncertainty level,
and $h(\mathcal{N}(\boldsymbol{\mu}, \boldsymbol{\sigma}^2))$ is the differential entropy of a multivariate Gaussian distribution which is actually only related to $\boldsymbol{\sigma}$:
\begin{equation}
    h(\mathcal{N}(\boldsymbol{\mu}, \boldsymbol{\sigma}^2)) = \frac{1}{2} \log (\det (2 \pi e \boldsymbol{\sigma}^2)).
\end{equation}
It is obvious that $L_{reg}$ will maintain the uncertainty level of the learned stochastic representations.

In conclusion, our final uncertainty-aware loss for the relationship model is expressed as:
\begin{equation}
    L_{rel} = L_{ce} + \alpha L_{reg},
\end{equation}
where $\alpha$ is the weight of regularization term.

\noindent \textbf{Reparameterization Trick.}
Sampling $\boldsymbol{z}$ from $\mathcal{N}(\boldsymbol{\mu}, \boldsymbol{\sigma^2})$ directly will cause the problem of preventing gradients from propagating back to the preceding layers.
Thus, we use the reparameterization trick~\cite{vae} to bypass the problem.
Specifically, we first sample a random noise $\epsilon$ from the standard Gaussian distribution and generate $\boldsymbol{z}$ as the equivalent sampling representation,
\begin{equation}
    \boldsymbol{z} = \boldsymbol{\mu} + \epsilon \boldsymbol{\sigma}, \epsilon \sim \mathcal{N}(\boldsymbol{0}, \boldsymbol{I}).
\end{equation}

\begin{table*}
\small
\setlength\tabcolsep{9pt}
\renewcommand{\arraystretch}{1.2}
\caption{Comparisons among various methods on mR@\emph{K} (\%).
$\dagger$ denotes the re-implemented version from \cite{motif}.
$\uparrow$ and $\downarrow$ indicate the performance change before and after plugging in PUM.
}
\begin{center}
\begin{tabular}{l|c|c|c|c|c|c|c}
\hline
& \multicolumn{2}{c|}{SGDet}
& \multicolumn{2}{c|}{SGCls}
& \multicolumn{2}{c|}{PredCls}
& \\
\hline\hline
Methods &
mR@50 & mR@100 &
mR@50 & mR@100 &
mR@50 & mR@100 &
Mean \\
\hline
IMP$\dagger$ ~\cite{imp,motif} & 3.8 & 4.8 & 5.8 & 6.0 & 9.8 & 10.5 & 6.8 \\
FREQ~\cite{motif} & 4.3 & 5.6 & 6.8 & 7.8 & 13.3 & 15.8 & 8.9 \\
SMN~\cite{motif} & 5.3 & 6.1 & 7.1 & 7.6 & 13.3 & 14.4 & 9.0 \\
KERN~\cite{kern} & 6.4 & 7.3 & 9.4 & 10.0 & 17.7 & 19.2 & 11.7 \\
VCTREE-SL~\cite{vctree} & 6.7 & 7.7 & 9.8 & 10.5 & 17.0 & 18.5 & 11.7 \\
VCTREE-HL~\cite{vctree} & 6.9 & 8.0 & 10.1 & 10.8 & 17.9 & 19.4 & 12.2\\
\hline
IMP$\dagger$ + PUM & 4.5\up{0.7} & 5.5\up{0.7} & 6.4\up{0.6} & 6.8\up{0.8} & 11.3\up{1.5} & 12.3\up{1.8} & 7.8\up{1.0} \\
SMN + PUM & 7.5\up{2.2} & 8.6\up{2.5} & 9.4\up{2.3} & 10.1\up{2.5} & 16.4\up{3.1} & 18.1\up{3.7} & 11.7\up{2.7} \\
KERN + PUM & 6.5\up{0.1} & 7.4\up{0.1} & 9.9\up{0.5} & 10.6\up{0.6} & 18.7\up{1.0} & 20.4\up{1.2} & 12.3\up{0.6} \\
VCTREE-SL + PUM & 7.1\up{0.4} & 8.2\up{0.5} & 11.0\up{1.2} & 11.9\up{1.4} & 19.0\up{2.0} & 20.9\up{2.4} & 13.0\up{1.3} \\
\hline
ResCAGCN~\cite{sgg_dual_resgcn} & \textbf{7.9} & 8.8 & 10.2 & 11.1 & 18.3 & 19.9 & 12.7 \\
\textbf{ResCAGCN + PUM} & 7.7\down{0.2} & \textbf{8.9}\up{0.1} & \textbf{11.9}\up{1.7} & \textbf{12.8}\up{1.7} & \textbf{20.2}\up{1.9} & \textbf{22.0}\up{2.1} & \textbf{13.9}\up{1.2} \\
\hline
\end{tabular}
\end{center}
\label{table:mean_recall}
\end{table*}

\begin{table}
\small
\setlength\tabcolsep{4.6pt}
\renewcommand{\arraystretch}{1.2}
\caption{Comparisons among various methods on R@100 (\%).
}
\begin{center}
\begin{tabular}{l|c|c|c}
\hline
& \multicolumn{1}{c|}{SGDet}
& \multicolumn{1}{c|}{SGCls}
& \multicolumn{1}{c}{PredCls}
\\
\hline\hline
Methods &
R@100 &
R@100 &
R@100 \\
\hline
IMP$\dagger$ ~\cite{imp,motif} & 24.5 & 35.4 & 61.3 \\
FREQ~\cite{motif} & 30.1 & 32.9 & 62.2 \\
SMN~\cite{motif} & 30.3 & 36.5 & 67.1 \\
KERN~\cite{kern} & 29.8 & 37.4 & 67.6 \\
VCTREE-SL~\cite{vctree} & 31.1 & 38.6 & 67.9 \\
VCTREE-HL~\cite{vctree} & 31.3 & 38.8 & 68.1 \\
\hline
IMP$\dagger$ + PUM & 25.0\up{0.5} & 35.7\up{0.3} & 61.8\up{0.5} \\
SMN + PUM & 30.6\up{0.3} & 37.4\up{0.9} & 67.5\up{0.4} \\
KERN + PUM & 29.8 & 37.1\down{0.3} & 67.5\down{0.1} \\
VCTREE-SL + PUM & 30.9\down{0.2} & 38.1\down{0.5} & 67.6\down{0.3} \\
\hline
ResCAGCN~\cite{sgg_dual_resgcn} & 30.9 & 38.7 & 67.9 \\
\textbf{ResCAGCN + PUM} & \textbf{31.3}\up{0.4} & \textbf{39.0}\up{0.3} & \textbf{68.3}\up{0.4} \\
\hline
\end{tabular}
\end{center}
\label{table:recall}
\end{table}

\section{Experiments}

\subsection{Experiment Setting}

\noindent \textbf{Dataset.}
We evaluate the proposed method on the popular large-scale Visual Genome (VG) benchmark~\cite{vg}, which originally contains 108,077 images with average annotations of 38 objects and 22 relationships per image.
Since the majority of the annotations are noisy, following previous works~\cite{motif,kern,vctree}, we adopt the most popular split from \cite{imp}, which selects top-150 object categories and top-50 predicate categories by frequency.

\noindent \textbf{Evaluation.}
We follow three conventional tasks to evaluate the proposed SGG model:
(1) \textbf{Predicate Classification (PredCls)}: given the bounding boxes and their object labels in an image, predict the predicates of all pairwise relationships.
(2) \textbf{Scene Graph Classification (SGCls)}: given the ground-truth bounding boxes in an image, predict the predicate as well as the object labels in every pairwise relationship.
(3) \textbf{Scene Graph Detection (SGDet)}: given merely an image, simultaneously detect a set of objects and predict the predicate between each pair of the detected objects.

Since the distribution of relationships in the VG dataset is highly imbalanced, we follow \cite{kern, vctree} to utilize mean Recall@\emph{K} (short as mR@\emph{K}) to evaluate each relationship in a balanced way.
For reference, all the methods are also evaluated by the conventional Recall@\emph{K} (short as R@\emph{K}) metric.

\noindent \textbf{Implementation Details.}
Following the prior works~\cite{motif,kern,vctree}, we adopt the same Faster-RCNN~\cite{faster_rcnn} to detect object bounding boxes and extract RoI features.
For the hyperparameters in PUM, we set $K$ to 8, $N$ to 8, $\lambda$ to 0.1, and $\gamma$ to 200.
We optimize the proposed model by the Adam optimizer with a batch size of 8, and momentums of 0.9 and 0.999.
Intuitively, our uncertainty modeling would cause variance of performances.
However, in practice, under the hyperparameter setting mentioned above, we observe that the variance is always negligible enough\footnote{0.03\% at most, with respect to mR/R@\emph{K}.} to be ignored.

\begin{figure}[ht]
    \centering\includegraphics[width=0.47\textwidth, keepaspectratio]{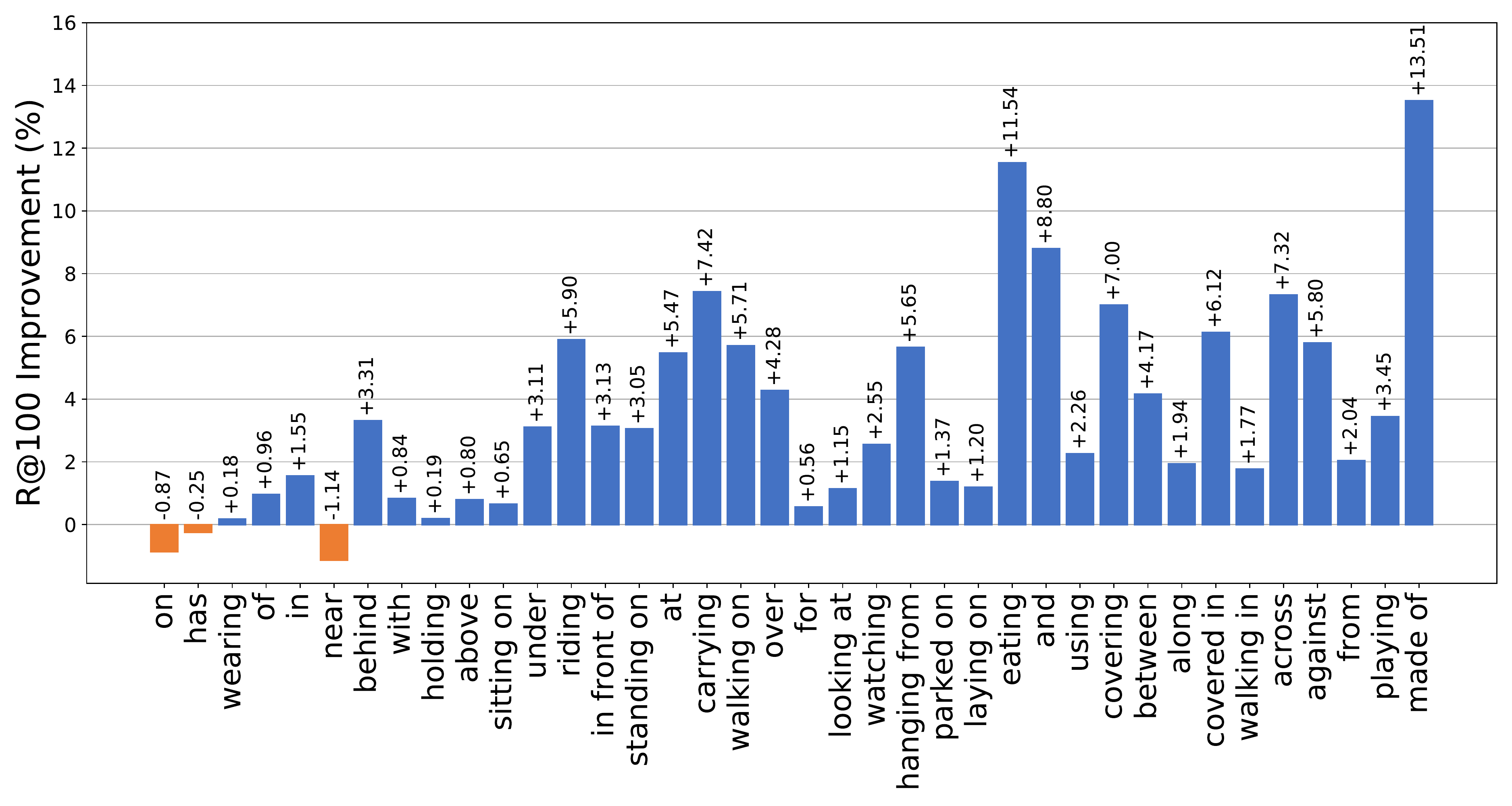}
	\caption{The R@100 improvement (\%) of different predicate categories over VCTREE-HL in the PredCls setting.
	The x-axis labels are in descending order according to their number of samples in the VG dataset.
	Categories with the same performances are filtered out.
	Blue (orange) color indicates increase (decrease) in performance.}
	\label{fig:pred_improv}
\end{figure}

\subsection{Comparisons with State-of-the-Art Methods}

\noindent \textbf{Comparing Methods.}
In this part, we compare our model with the existing state-of-the-art methods:
(1) designed to merely improve the recall metric, including Iterative Message Passing (\textbf{IMP})~\cite{imp}, frequency baseline without using visual contexts (\textbf{FREQ})~\cite{motif} and Stacked Motif Network (\textbf{SMN})~\cite{motif}; (2) intended for more balanced prediction on relationships, including Knowledge-Embedded Routing Network (\textbf{KERN})~\cite{kern} and Visual Context Tree model (\textbf{VCTREE-SL}, trained by supervised learning, and \textbf{VCTREE-HL}, trained by hybrid learning)~\cite{vctree}.
Although \cite{sgg_gps} also reported new state-of-the-art performances recently, we argue that their results are not directly comparable to ours.
Please refer to the supplementary material for details.

\noindent \textbf{Quantitative Results.}
From Table \ref{table:mean_recall}, compared with the previous state-of-the-art methods, the proposed model (\textbf{ResCAGCN + PUM}) shows the best performances on the mR@\emph{K} metric, with a relative improvement of 13.9\% compared with VCTREE-HL according to the mean.
This indicates that the proposed model achieves notable improvement on infrequent predicate categories.
Meanwhile, it does not sacrifice frequent categories a lot, since its performances on R@100 also reach state-of-the-art, as demonstrated in Table \ref{table:recall}.

To gain a more comprehensive understanding of this phenomenon, as depicted in Figure \ref{fig:pred_improv}, we further present the R@100 improvement of each predicate category over VCTREE-HL in the PredCls setting.
Note that the x-axis labels are in descending
order according to their number of samples in the VG dataset and categories with the same performances are filtered out.
It is obvious that the proposed model achieves significant improvement in most categories.
Importantly, the improvement is much larger for those infrequent categories in the long tail.
We mainly owe this phenomenon to the byproduct of PUM, which endows the model with more chances to cover infrequent categories and thus alleviates the issue of biased predictions.

\subsection{Ablation Study}

To better prove the effectiveness of PUM, in this part, we explore the gain of PUM as a plug-and-play module.
Firstly, as illustrated in the third part of Table \ref{table:mean_recall}, PUM brings about a significant increase over the vanilla ResCAGCN~\cite{sgg_dual_resgcn} by a margin of 1.2\% on mR@\emph{K}, according to the mean.
Meanwhile, from Table \ref{table:recall}, PUM also improves the model moderately on R@\emph{K}.
The results suggest that PUM indeed plays a critical role in the proposed model, which especially lies in more balanced predictions.

Then, we apply PUM to the existing state-of-the-art methods, which conventionally utilize deterministic representations for visual relationships.
Specifically, by regarding the original relationship features in each model as the $\eij$ in Eq. \ref{eq:rel_feature}, we adopt exactly the subsequent operations in Section \ref{sec:uncer_modeling} to model the uncertainty of relationships in a probabilistic manner.
We present comparisons on mR@\emph{K} between the existing state-of-the-art methods (IMP~\cite{imp}, SMN~\cite{motif}, KERN~\cite{kern} and VCTREE-SL\footnote{Since the reinforcement learning of VCTREE is independent of our method, we only conduct experiments on its one-stage supervised version for simplicity.}~\cite{vctree}) and the counterparts plugged in PUM in the second part of Table \ref{table:mean_recall}. 
We find that PUM improves the performances of all models by a significant margin, with a relative improvement of 14.7\%, 30.0\%, 5.1\% and 11.1\% compared with the baselines respectively.
The results show the universal superiority of PUM to the conventional deterministic modeling, which mainly lies in the effectiveness to alleviate the issue of biased predictions towards frequent relationships.
We also present comparisons on R@\emph{K} in Table \ref{table:recall}.
Note that PUM does not necessarily improve all baselines under this metric.
However, according to the discussions by Tang \etal~\cite{sgg_unbiased}, R@\emph{K} is not a ``qualified'' metric for SGG, since simply catering to frequent categories while neglecting the infrequent ones would unfairly obtain a good performance.
Meanwhile, similar to \cite{sgg_unbiased}, we also observe that the performance drops caused by PUM mainly originate from classifying trivial ``head'' predicates into more fine-grained ``tail'' classes, \eg from \texttt{on} to \texttt{parked on}.

\subsection{Understand Uncertainty Modeling} \label{sec:understand_uncer}

We observe that the proposed model can generate diverse relationships inherently, which helps to address the \emph{implicit multi-label} issue caused by the semantic ambiguity.
In this part, we qualitatively and quantitatively analyze this characteristic to gain more insights about our uncertainty modeling.

\begin{figure*}[h]
    \centering\includegraphics[width=\textwidth, keepaspectratio]{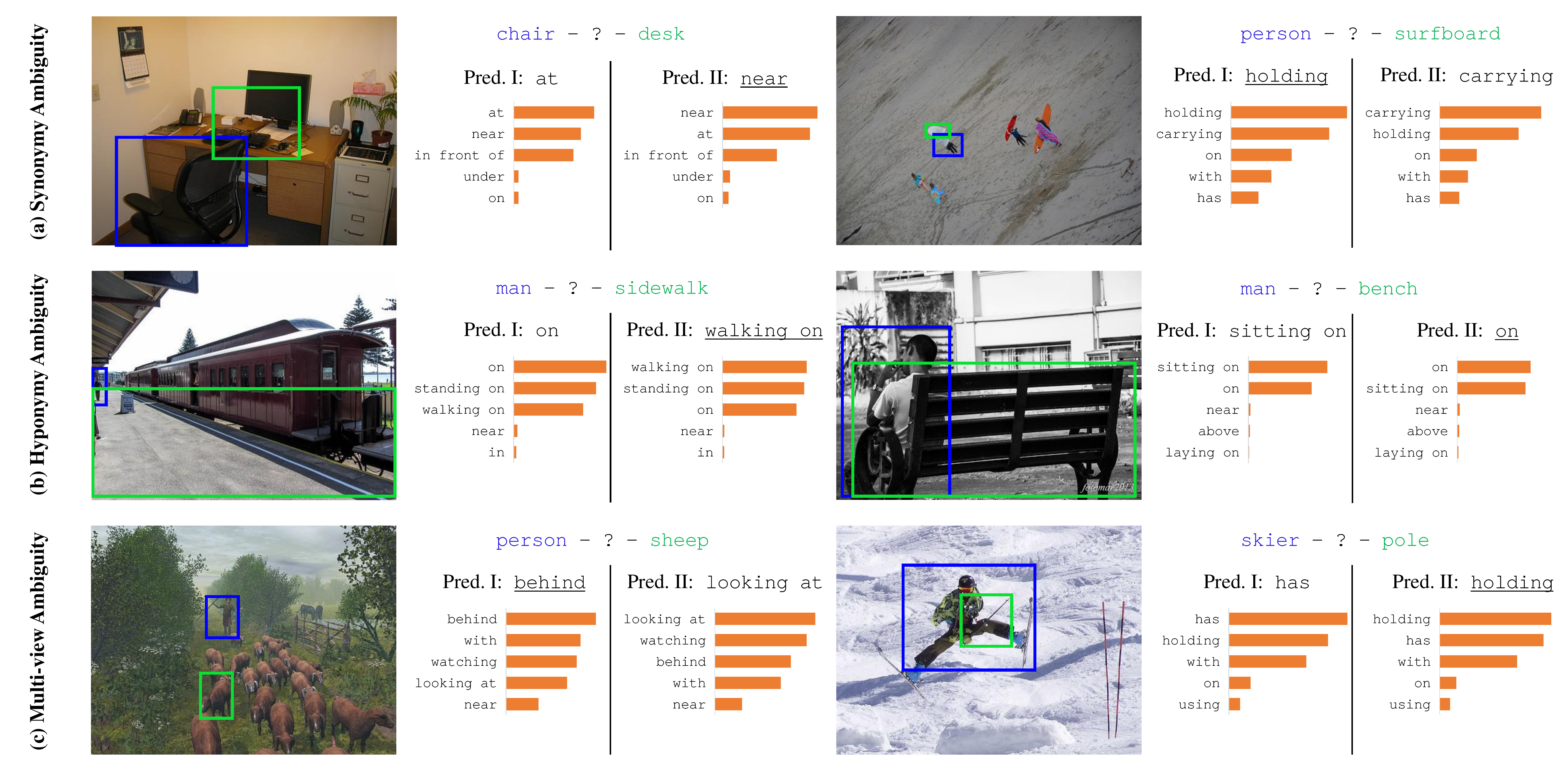}
	\caption{Qualitative examples from two consecutive predictions of the proposed model in the PredCls setting.
	Blue color indicates subjects; green color indicates objects; underlining indicates predictions that hit the ground-truth.
	For each prediction, we also show the top-5 classes with the highest confidences.
	Note that the confidences vary between two consecutive predictions, because of the stochasticity of the relationship features.}
	\label{fig:multi_infer}
\end{figure*}

\noindent \textbf{Qualitative Results.}
From Eq. \ref{eq:rel_cls}, the relationship features fed into the classifier are randomly drawn from Gaussian distributions, resulting in varied predicted confidences, even for the same union region.
Therefore, given a pair of objects, the proposed model can produce different plausible predicate at each inference.
In other words, it is able to describe the same visual scene in different ways, leading to more human-like diverse predictions.
This diversity well matches the three types of ambiguity illustrated in Figure \ref{fig:ambiguous_examples}.
We show qualitative examples from two consecutive predictions of the proposed model in the PredCls setting in Figure \ref{fig:multi_infer}.
From the first row, the proposed model generates semantically-similar predicates consecutively, \ie \texttt{at} vs. \texttt{near} and \texttt{holding} vs. \texttt{carrying}.
Although the ground-truth only considers a single label, we argue that there exactly exists such Synonymy Ambiguity, where multiple synonyms are plausible at the same time.
Hyponymy Ambiguity is also a common phenomenon, where predicates across adjacent abstract levels are interchangeable.
In the second row, the ground truth can be fine-grained (\texttt{walking on}) or coarse-grained (\texttt{on}).
Thanks to our uncertainty modeling, the proposed model covers both levels of granularity and thus increases the chance of hitting the ground-truth.
In Figure \ref{fig:ambiguous_examples} (c), different human annotators tend to describe similar visual scenes from different points of view, resulting in Multi-view Ambiguity.
We observe that the proposed model also simulates this phenomenon well.
From Figure \ref{fig:multi_infer} (c), for the relationship between \texttt{person} and \texttt{sheep}, the proposed model focuses on either the spatial position (\texttt{behind}) or the person's action (\texttt{looking at}).
On the other hand, for the scene on the right, the predictions could be a possessive verb (\texttt{has}) or an actional verb (\texttt{holding}).
In a word, these qualitative examples suggest that semantic ambiguity is quite common when describing visual relationships, while the proposed model manages to generate \emph{diverse} yet \emph{plausible} predictions. 

\noindent \textbf{Oracle Evaluation.}
Inspired by the \emph{oracle error rate} in Multiple Choice Learning~\cite{mcl, smcl, cmcl, vmcl}, we propose to use oracle Recall (short as oR) to measure the diversity of predictions indirectly, which counts a hit if one of the multiple consecutive predictions matches the ground-truth and can be formulated as follows:

\begin{equation} \label{eq:oR}
    oR = \frac{\sum^R_{i=1} \mathbf{1}(\sum^M_{m=1} \mathbf{1}(\hat{y}_{m,i} = y_i) > 0)} {R},
\end{equation}

\begin{equation}
    \mathbf{1}(x) = 
    \left\{\begin{array}{ll}
    1 & x = \text{True,} \\
    0 & x = \text{False.}
    \end{array}\right.
\end{equation}
Above, $R$ is the number of the ground-truth relationships in an image, $M$ is the number of consecutive predictions, $\mathbf{1}(\cdot)$ is an indicator function, and $\hat{y}_{m,i} = y_i$ means the $m$-th prediction on the $i$-th relationship hits the ground-truth.
if $M$ is set to 1, then oR is reduced to the normal recall metric.
Note that we omit the averaging over all images in Eq. \ref{eq:oR} for clarity.

\begin{figure}[t]
    \centering\includegraphics[width=0.45\textwidth, keepaspectratio]{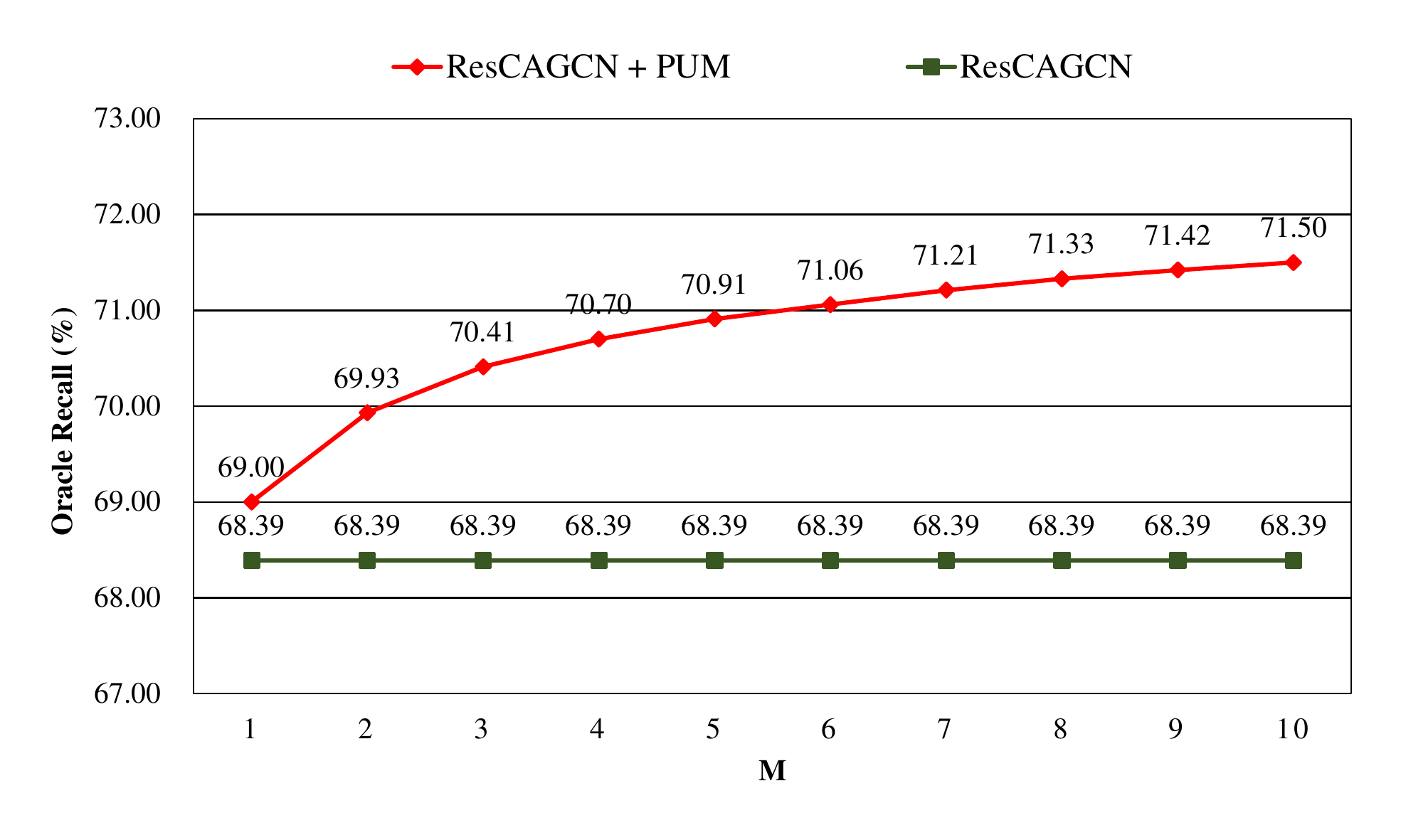}
	\caption{The effects of different number of consecutive predictions ($M$) on oR (\%).
	The performance of ResCAGCN + PUM is higher than that of ResCAGCN and increases with $M$.}
	\label{fig:oR}
\end{figure}

In order to focus on the prediction of predicates, we only conduct experiments in the PredCls setting.
It is obvious that a model with the ability to make inferences diversely will achieve better performance under this metric.
As illustrated in Figure \ref{fig:oR}, we evaluate the oR of the proposed model with and without PUM, respectively.
As $M$ increases, while the performance of ResCAGCN remains unchanged due to the lack of diversity, ResCAGCN + PUM gets improved steadily.
The result suggests that our uncertainty modeling not only boosts the coverage of predicted relationships in a single inference (as indicated when $M = 1$), but generates fresh relationships diversely in the next consecutive new predictions, which improves the opportunities to hit the ground-truth.

\section{Conclusion}

In this work, we considered the semantic ambiguity of visual relationships, which could be classified into Synonymy Ambiguity, Hyponymy Ambiguity and Multi-view Ambiguity.
To address the implicit multi-label issue caused by the ambiguity, we proposed a novel plug-and-play module dubbed PUM.
Although we were meant for diverse predictions, thanks to the byproduct of PUM, we achieved state-of-the-art performances under the existing evaluation metrics when combining it with ResCAGCN.
Furthermore, we proved the universal effectiveness of PUM and explored its ability to generate diverse yet plausible relationships both qualitatively and quantitatively.
A possible future direction would be to apply this kind of uncertainty modeling in down-stream tasks that also emphasize diversity, such as diverse visual captioning~\cite{img_cap_diverse, img_cap_adv, video_cap_diverse}.

\section*{Acknowledgements}

This research was partially supported by the Key Program of National Natural Science Foundation of China under Grant No. U1903213, the Dedicated Fund for Promoting High-Quality Economic Development in Guangdong Province (Marine Economic Development Project: GDOE[2019]A45), and the Guangdong Basic and Applied Basic Research Foundation (No. 2019A1515011387).
Baoyuan Wu is supported by the Natural Science Foundation of China under grant No. 62076213, the university development fund of the Chinese University of Hong Kong, Shenzhen under grant No. 01001810, and the special project fund of Shenzhen Research Institute of Big Data under grant No. T00120210003.

{\small
\bibliographystyle{ieee_fullname}
\bibliography{main}
}


\end{document}